\documentclass[twoside,11pt]{article}

\usepackage{CJKutf8} 
\usepackage{bxcjkjatype} 
\usepackage[T1]{fontenc}
\usepackage[utf8]{inputenc}

\usepackage[a4paper]{geometry}
\usepackage{RR}

\usepackage[unicode,colorlinks,allcolors=black]{hyperref} 
\usepackage[inline]{enumitem}

\usepackage[titles]{tocloft}
\usepackage{xcolor}
\usepackage{csquotes}

\usepackage{tabularx}
\usepackage{url}

\usepackage{nameref}

\usepackage{mathtools}
\usepackage{amsmath}
\usepackage{amsfonts}
\usepackage{calrsfs}
\DeclareMathAlphabet{\pazocal}{OMS}{zplm}{m}{n}

\usepackage{amssymb}
\usepackage{amsthm}

\newtheorem{example}{Example}

\usepackage[french,british]{babel}

\usepackage{pgf}
\usepackage{tikz}
\usetikzlibrary{shapes}
\usetikzlibrary{fit}
\usetikzlibrary{calc}
\usetikzlibrary{decorations.pathmorphing}
\usetikzlibrary{decorations.markings}
\usetikzlibrary{patterns}
\usetikzlibrary{patterns.meta}

\usepackage[outputdir={dotfiles/}]{dot2texi}
\tikzstyle{concept} = [
        align=center,
        draw,
        rectangle split,
        rectangle split parts=2,
        rounded corners=.2cm,
        minimum width=1cm,
        minimum height=1cm,
      ]
\tikzstyle{smconcept} = [
        align=center,
        draw,
        rectangle split,
        rectangle split parts=2,
        rounded corners=.2cm,
        minimum width=1cm,
        minimum height=1cm,
      ]
\tikzstyle{nosep} = [ inner sep=0pt, outer sep=0pt ]

\usepackage[
        backend=biber,
	backref=true,               
	url=true,                      
        doi=true,
 	maxcitenames=2,         
 	maxbibnames=100,     
        style=apa,
 	sorting=nyt,                    
 	hyperref=true                
 	]{biblatex} 

\DeclareFieldFormat{bibhyperrefnonest}{%
  \DeclareFieldFormat{bibhyperref}{##1}%
  \bibhyperref{#1}}

\DeclareFieldFormat{doi}{%
   \mkbibacro{DOI}\addcolon\space 
   \ifhyperref 
     {\href{https://doi.org/#1}{\nolinkurl{#1}}} 
     {\footnotesize\nolinkurl{#1}}}

\DeclareFieldFormat{url}{%
   \ifhyperref 
     {\url{#1}}
     {\footnotesize\nolinkurl{#1}}}

\DeclareCiteCommand{\cite}[\mkbibbrackets]
  {\usebibmacro{cite:init}\usebibmacro{prenote}}
  {\usebibmacro{citeindex}\printtext[bibhyperrefnonest]{\usebibmacro{cite}}}
  {}
  {\usebibmacro{postnote}\usebibmacro{cite:post}}

\AtEveryBibitem{\clearfield{note}}
\AtEveryBibitem{\clearfield{language}}

\setlist{leftmargin=.5cm,topsep=0pt,parsep=-5pt}
\setlist[2]{itemsep=4pt}
\setlist[description]{font={\normalfont\sffamily}}

\newlist{enuminline}{enumerate*}{5}
\setlist[enuminline]{label={(\arabic*)}}

\newlist{enumii}{enumerate*}{5}
\setlist[enumii]{label={(\emph{\roman*})}}

\newlist{enumaa}{enumerate*}{5}
\setlist[enumaa]{label={(\emph{\alph*})}}

\let\emptyset\varnothing

\newcommand\fullversion[2]{#2}

\usepackage{bm}
\usepackage{multirow}
\usepackage{diagbox}
\usepackage{graphicx}
\usepackage{interval}
\usepackage{stfloats}

\usepackage[table]{xcolor}

\usepackage{pgfplots}
\usepackage{pgfplotstable}
\usepgfplotslibrary{statistics}
\pgfplotsset{compat=1.16}
 \pgfplotsset{
    only if/.style args={entry of #1 is #2}{
        /pgfplots/boxplot/data filter/.code={
            \edef\tempa{\thisrow{#1}}
            \edef\tempb{#2}
            \ifx\tempa\tempb
            \else
                
            \fi
        }
    }
} 

\usepackage{tikz}
\usetikzlibrary{arrows}
\usetikzlibrary{fit}
\usetikzlibrary{trees}
\usetikzlibrary{calc}
\usetikzlibrary{positioning}
\usetikzlibrary{shapes.callouts}
\usetikzlibrary{pgfplots.statistics}

\tikzset{
echo chamber/.style={blue, thick, rounded corners},
component/.style={blue, thick, rounded corners,dashed},
}

\usepackage{amsmath}
\usepackage{mathtools}
\usepackage{booktabs}
\usepackage{comment}

\usepackage{enumitem}
\newlist{enumerateA}{enumerate}{5}
\setlist[enumerateA]{
  leftmargin = *,
  topsep=0pt,
  align = left,
  label =(\Alph*),
  labelwidth=\parindent,
  labelsep = 2pt,
  itemindent = 0pt
}

\newcommand\xusecounter[2]{\refstepcounter{#1}\label{#2}(\alph{#1})}
\newcommand\xfigref[2]{Figure~\ref{#1} (\ref{#2})}

\newcommand{\beliefrevision}{\ensuremath{BR}}
\newcommand{\opinionform}{\ensuremath{OF}}
\newcommand{\opiniondynamics}{\ensuremath{OD}}
\newcommand{\beliefalign}{\ensuremath{BA}}

\usepackage{todonotes} 
\newcounter{jecomment}

\newcounter{hirocomment}

\newcounter{kojicomment}

\newcommand{\hiroForLater}[1]{}

\newcommand\uppercasing[2]{#1}

\DeclareMathOperator*{\argmin}{argmin}

\newcommand{\measure}[1]{\textsf{\textbf{#1}}}
\newcommand{\hypothesis}[1]{\textbf{H#1}}

\newcommand{\sccs}[1][t]{\mathcal S^{#1}} 

\renewenvironment{example}{}{}

\RRNo{9635}
\RRdate{November 2026}
\RRauthor{Hiro Kataoka\thanks{\begin{tabular}{rl}\\
Affiliation:& University of Tsukuba, Tsukuba, Japan\\
Contact:& Hiro.Kataoka@mas.cs.tsukuba.ac.jp \end{tabular}} \and Jérôme Euzenat\thanks{\begin{tabular}{rl}\\
Affiliation:& Univ. Grenoble Alpes, Inria, CNRS, Grenoble INP, LIG, F-38000 Grenoble, France\\
Contact:& Jerome.Euzenat@inria.fr\end{tabular}} \and Koji Hasebe\thanks{\begin{tabular}{rl}\\
Affiliation:& University of Tsukuba, Tsukuba, Japan\\
Contact:& hasebe@cs.tsukuba.ac.jp \end{tabular}}}
\RRtitle{La propagation de croyances et d'opinions cohérentes produit plus de chambres d'écho:\\
version corrigée}
\RRetitle{Coherent belief and opinion propagation produces more echo chambers:\\
corrected version}
\RRjtitle{\begin{CJK}{UTF8}{ipxg}一貫した意見と信念の伝播が\\エコーチェンバーの形成に与える影響の分析（改訂版）\end{CJK}}
\titlehead{Coherent belief and opinion propagation produces more echo chambers}
\authorhead{Hiro Kataoka, Jérôme Euzenat, Koji Hasebe}

\RRresume{
  \begin{otherlanguage}{french}
    Le développement de chambres d'écho dans les réseaux sociaux est un phénomène bien connu.
    Les chambres d'écho sont des groupes d'agents satisfaisant trois propriétés:
    homogénéité, ils partagent la même opinion,
    ségrégation, ils sont connectés principalement à des agents du groupe, et
    renforcement, leur homogénéité augmente au cours du temps.
    La dynamique des opinions a été utilisée pour expliquer l'émergence de chambres d'écho dans des population d'agents partageant leurs opinions.
    Cependant, dans les réseaux sociaux, les agents échangent aussi leurs croyances expliquant leurs opinions.
    Dans ce rapport, nous cherchons à caractériser l'effet de la propagation de croyances avec les opinions sur la formation des chambres d'écho.
    À cet effet, on utilise en complément d'un modèle de dynamique des opinions, un modèle de propagation des croyances.
    La cohérence des croyances et des opinions est maintenue par les valeurs que les agents partagent.
    On introduit une mesure, s'appliquant indépendamment aux croyances et aux opinions, pour identifier les chambres d'écho sur la base des trois propriétés mentionnées.
    Ce modèle est utilisé pour simuler la propagation d'opinions et de croyances en fonction de différentes conditions initiales et de différentes tolérances des agents pour les opinions et croyances différentes.
    Les résultats confirment la présence de chambres d'écho observée dans des études précédentes avec un modèle différent et une mesure plus stricte.
    Ils montrent aussi la présence de chambre d'écho dans le cadre de la propagation de croyances.
    Enfin, ils montrent que la propagation jointe des opinions et des croyances, d'une part, et le maintien de leur cohérence augmente le nombre de chambres d'écho basées sur les croyances, mais qu'ils impliquent un nombre plus réduit d'agents.
    
\end{otherlanguage}
}

\RRjabstract{
\begin{CJK}{UTF8}{ipxm}
エコーチェンバーの形成は，ソーシャルネットワークにおいて観察されるよく知られた現象である．%
エコーチェンバーとは，同じ意見を共有する「同質性」，主に同じ集団のエージェントと相互作用する「分離性」，時間とともに意見の類似度が高まる「強化」という三つの性質を満たすエージェントの集団を指す．
これまで，エージェントが意見を更新することを通してエコーチェンバーが形成される過程を説明するために，様々な意見形成モデルが提案されてきた．
しかし，ソーシャルネットワークのユーザは，自らの意見だけでなくそれを裏づける信念も他のユーザに伝えている．
そこで本研究では，意見とともに信念が伝播することがエコーチェンバーの形成に与える影響を明らかにすることを目的とする．
そのために，既存の意見形成モデルに古典的な信念改訂操作を組み込んだ上で，エージェントが共有する価値観に基づいて信念と意見の間の一貫性が保たれるよう，先行研究で提案されたモデルを拡張する．
さらに，エコーチェンバーの三つの性質に基づいて，意見または信念に関するエコーチェンバーを特定する尺度を提案する．
このモデルを用いて，初期条件と，異なる意見や信念に対するエージェントの許容度を変化させながら，意見と信念の伝播をシミュレーションした．
その結果，先行研究とは異なる設定において，より厳格な尺度のもとで，先行研究と同様にエージェントが意見に関するエコーチェンバーを形成することが確認された．
また，信念の伝播のみを行う場合でも，信念に関するエコーチェンバーが形成することが示された．
最後に，意見を信念と結びつけて両者の一貫性を維持するとき，信念に関するエコーチェンバーの数は増えるが，それぞれに含まれるエージェントの数は減少することが明らかになった．
\end{CJK}
}

\RRabstract{
The occurrence of echo chambers is a well-known phenomenon in social networks.
Echo chambers are groups of agents satisfying three properties: homogeneity, i.e., sharing the same opinion, segregation, i.e., listening primarily to agents of the same group, and reinforcement, i.e., increasing opinion similarity over time. Opinion dynamics models have been used to explain how echo chambers occur through agents updating their opinions.
However, social network users also exchange beliefs supporting their opinions.
Here, we aim at identifying the effect of propagating beliefs together with opinions on the formation of echo chambers.
For that purpose, we complement an existing opinion dynamics model with classical belief merge operations.
Beliefs and opinions are also maintained coherent guided by values that agents share. 
We introduce measures, applying to either beliefs and opinions, to identify echo chambers based on the three properties.
This model is used to simulate opinion and belief propagation with varying initial conditions and agent tolerance for different opinions/beliefs.
Results confirm the occurrence of echo chambers observed in previous opinion dynamics studies in a different setting and with a stricter measure.
They also demonstrate the occurrence of echo chambers with belief propagation.
Finally, they show that connecting opinions to beliefs and maintaining their coherence increases the number of \textcolor{blue}{belief}
echo chambers, but that they involve a reduced number of agents.
}

\RRmotcle{Chambre d'écho; Dynamique des opinions; Révision des croyances; Cohérence fondée sur les valeurs; Simulation multi-agents}
\RRkeyword{Echo chambers; Opinion dynamics; Belief revision; Value-based coherence; Multi-agent simulation}
\RRjkeywords{\begin{CJK}{UTF8}{ipxm}エコーチェンバー; 意見形成; 信念改訂; 価値観に基づく一貫性; マルチエージェントシミュレーション\end{CJK}} 

\RRprojet{mOeX\\
  Fukuro Associate Team}
\URRhoneAlpes
\RCGrenoble

\newenvironment{starfootnotes}
  {\par\edef\savedfootnotenumber{\number\value{footnote}}
   \setcounter{footnote}{0}}
 {\par\setcounter{footnote}{\savedfootnotenumber}}

\begin{document}

\begin{starfootnotes}
\makeRR
\end{starfootnotes}

\setcounter{tocdepth}{1}

\renewcommand{\baselinestretch}{0.95}\normalsize
\tableofcontents
\renewcommand{\baselinestretch}{1.0}\normalsize

\bigskip
{\color{blue}\hypersetup{allcolors=blue}
We discovered a mistake in the code for analyzing the results of \cite{kataoka2026c}.
This mistake affects the way the reinforcement measure is computed, and ultimately how echo chambers are identified.
It entails that all echo chamber figures reported in the paper were incorrect.
However, they still supports 5 out of the 6 tested hypotheses.
The remaining hypothesis is not supported:
The number of opinion echo chambers is \emph{not} increased by synchronization over joint propagation of opinions and beliefs.

This report is a version of \cite{kataoka2026c} whose measures have been corrected.
The corrections are displayed in blue.
\hypersetup{allcolors=black}}

\section{Introduction}

Echo chambers are well-known phenomena in social networks.
They denote a situation in which a group of persons only interacts with people who have opinions close to theirs, thus reinforcing them \cite{Mahmoudi2024}.
They have unwanted effects in terms of social cohesion.

Opinion dynamics models have been applied to explain how echo chambers arise by aligning agents' opinions with those that they receive from others.
For example, it is shown that, similarly to existing social networks, echo chambers in online social networks are created through ignoring diverging opinions and modifying the network to avoid them~\cite{Sasahara2020}.

However, these models do not take into account the effect of agents' beliefs:
people exchange and update their beliefs as well as their opinions through interactions.
For example, beliefs can be communicated to justify opinions:
\begin{example}
``\emph{I am for constructing tram lines} (opinion) because \emph{I believe that this would reduce traffic jams} (justification by beliefs).''
\end{example}
Hereafter, opinions and beliefs are different: the opinions of an agent denote how positive or negative it is toward a topic while beliefs represent what it considers as true based on logical reasoning.
The way in which beliefs can propagate has been studied in the field of
belief change \cite{ferme2018a}.

Based on these observations, 
we aim at investigating the effect of introducing beliefs on echo chambers:
Does it increase or decrease the number of opinion echo chambers?
Do belief and opinion echo chambers coincide?
To achieve this objective, 
we introduce a model defining interactions between opinions and beliefs.
In this model, opinions and beliefs are propagated through existing models \cite{Sasahara2020,Schwind_Inoue_Bourgne_Konieczny_Marquis_2015} and a value-based mechanism is used to maintain their coherence \cite{kataoka2025a,kataoka2026a}.
Using the proposed model, we aim at studying whether allowing agents to interact based on beliefs leads to more or less echo chambers.

Various measures have been provided to identify echo chambers in online social networks and simulations~\cite{Interian_GMarzo_Mendoza_Ribeiro_2023,Clauset_Newman_Moore_2004,Mahmoudi2024,Hartmann_Wang_Pohlmann_Berendt_2025}.
However, such measures do not necessarily cover all of the characteristics of echo chambers.
Thus, we introduce measures to capture echo chambers based on their definition in existing work.
Introducing these measures allows us to provide new and more precise results about the influence of belief introduction on echo chambers.

Experiments confirm that agents form echo chambers either based on their opinions \cite{Sasahara2020} or their beliefs \cite{kataoka2025a}.
In addition, they show that performing belief and opinion propagation without maintaining their coherence already generates more echo chambers.
Finally, ensuring coherence between agents' opinions and beliefs leads to even more, albeit smaller, \textcolor{blue}{belief} echo chambers.

The remainder of this paper is organized as follows.
Section~\ref{sec:preliminaries} introduces notations used in this paper.
Section~\ref{sec:related-work} presents work related to this problem.
The new integrated model is defined in Section~\ref{sec:model}.
Section~\ref{sec:measures} defines a measure to identify echo chambers and
Section~\ref{sec:hypotheses} draws hypotheses to be tested in this paper.
Then, the experimental settings are described in Section~\ref{sec:experimental-settings}.
Section~\ref{sec:results} presents and discusses the experimental results.

\section{Preliminaries}\label{sec:preliminaries}

We introduce
concepts relevant to both related work and ours.

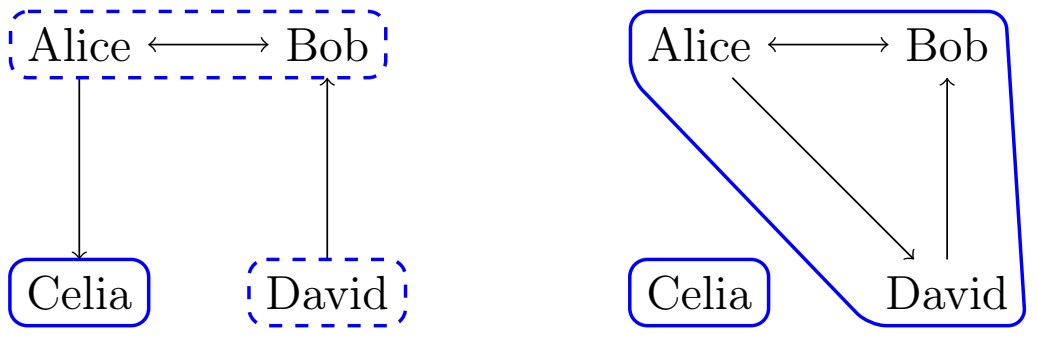
\begin{figure}
 \centering
  \scalebox{.82}{
 \begin{tikzpicture}
  \node (alice) at (0, 0) {{Alice}};
  \node (bob) at (2, 0) {{Bob}};
  \node (charles) at (0, -2) {{Celia}};
  \node (david) at (2, -2) {David};
  \draw [<->] (alice) -- (bob);
  \draw [->] (alice) -- (charles);
  \draw [->] (david) -- (bob);
  \draw[component] (alice.north west) -- (bob.north east) -- (bob.south east) -- (alice.south west) -- cycle;
  \draw[echo chamber] (charles.north west) -- (charles.north east) -- (charles.south east) -- (charles.south west) -- cycle;
  \draw[component] (david.north west) -- (david.north east) -- (david.south east) -- (david.south west) -- cycle;

  \begin{scope}[xshift=5cm]
  \node (alice) at (0, 0) {{Alice}};
  \node (bob) at (2, 0) {{Bob}};
  \node (charles) at (0, -2) {{Celia}};
  \node (david) at (2, -2) {David};
  \draw [<->] (alice) -- (bob);
  \draw [->] (alice) -- (david);
  \draw [->] (david) -- (bob);

  \draw[echo chamber] (alice.north west) -- (bob.north east) -- (david.south east) -- (david.south west) -- (alice.south west) -- cycle;
  \draw[echo chamber] (charles.north west) -- (charles.north east) -- (charles.south east) -- (charles.south west) -- cycle;
 \end{scope}
 \end{tikzpicture}}
 \caption{Visualization of a sample network before (left) and after (right) rewiring.
An arrow from a person $p$ to another person $q$ means that 
$p$ has access to $q$'s opinions and beliefs.
Blue dashed lines circumscribe strongly connected components; blue solid lines are echo chambers.}
\label{fig:intuitive-scenario-network}
\end{figure}

Let $A$ be a set of agents.
Each agent $a\in A$ has a non-empty set of neighbors $N_a^t\subseteq A\setminus\{a\}$ at time $t$.
The directed network that agents form at time $t$ is denoted as $G^t=\langle A,N^t\rangle$ such that $N^t=\bigcup_{a\in A}\{\langle a,a'\rangle;a'\in N_a^t\}$.
\begin{example}
For the sake of example, one can consider a set of agents called Alice, Bob, Celia, and David.
The network they form is visualized in Figure~\ref{fig:intuitive-scenario-network} (left).
Here, Bob and Celia are Alice's neighbors, i.e., $N_\text{Alice}^0=\{\text{Bob},\text{Celia}\}$.
\end{example}

Each agent $a\in A$ also has a cognitive state $\langle O^t_a, B^t_a\rangle$ at time $t$ made of its opinions $O_a^t$ and beliefs $B_a^t$.

An agent's beliefs are a set of formulas held true by the agent.
Here we consider agents' beliefs $B$ as sets of formulas in a propositional logic language $\mathcal{L_P}$, built from a set $\mathcal P$ of propositions.
\begin{example}
Such propositions may be that `a tram be built' (noted as $p$), `traffic jams are low' ($q$) or `taxes are low' ($r$).
If Alice believes that trams reduce traffic jams, her beliefs may be $B_\text{Alice}^0=\{p\rightarrow q\}$.
Similarly, Bob may believe that
`building tram lines increases taxes ($B_\text{Bob}^0=\{p\rightarrow \neg r\}$)' and Celia that if taxes should be kept low, then trams cannot be built ($B_\text{Celia}^0=\{r\rightarrow\neg p\}$).
\end{example}

The belief set $B$ can always be replaced by a single formula: the conjunction of its elements.
In this work, this set of beliefs is expressed, without loss of generality, as a single formula $[B]$ in full disjunctive normal form (FDNF), i.e. each literal appearing exactly once in each disjunct.
\begin{example}
Hence, Alice's beliefs are expressed as $(\neg p\wedge \neg q\wedge r)\vee(\neg p\wedge q\wedge r)\vee(\neg p\wedge \neg q\wedge \neg r)\vee(\neg p\wedge q\wedge \neg r)\vee(p\wedge q\wedge \neg r)\vee(p\wedge q\wedge r)$.
\end{example}

The set of models of a belief set $B\subseteq\mathcal{L_P}$ is denoted by $\mathcal{M}(B)$.
There is a one-to-one correspondence between the disjuncts of the formula $[B]$ and the models of the belief set $\mathcal{M}(B)$.
The symbol $\top$ (top) denotes the formula satisfied in by all interpretations (or the empty theory), thus $\mathcal{M}(\top)$ is the set of all possible interpretations.

Opinions are represented as numbers between $0$ and $1$, i.e., $O_a^t\in\interval{0}{1}\subseteq\mathbb{R}$ with respect to topics.
In this paper, we consider a single topic $\phi$ which is a formula, i.e., $\phi\in\mathcal{L_P}$.
\begin{example}
For instance, Alice, Bob, Celia, and David may discuss whether the city they are living in should construct tram lines or not.
Hence they have opinions about the topic $\phi=p$ (`a tram be built').
Initially, Alice is in favor of the construction plan ($O_\text{Alice}^0=1.0$), while Bob and Celia are less favorable ($O_\text{Bob}^0=0.6$ and $O_\text{Celia}^0=0.4$).
\end{example}
Intuitively, opinions $0$ and $1$ represent extremely negative and positive opinions, respectively.

The distance between two opinions $O$ and $O'$ is measured by the absolute difference between their values:
\begin{align}
 d_O(O,O')&=|O-O'|.
 \label{eq:do}
\end{align}
Similarly, the distance between two belief sets $B$ and $B'$ is defined as the Hamming distance between their models:
\begin{align}
  d_B(B,B')&=|\mathcal{M}(B)\setminus\mathcal{M}(B')|+|\mathcal{M}(B')\setminus\mathcal{M}(B)|.
 \label{eq:db}
\end{align}

Agents receive opinions and beliefs from their neighbors only.
At each time $t$, agent $a$ can determine its set of concordant neighbors $C_a^t$ as those neighbors having close opinions and beliefs by:
\[
 C_a^t=\{a'\in N_a^t;d_O(O_a^t,O_{a'}^t)\leq\varepsilon\land d_B(B_a^t,B_{a'}^t)\leq\delta\}
\]
\noindent such that $\varepsilon\in\interval{0}{1}$ and $\delta\in\{0,1,\ldots,|\mathcal{M}(\top)|\}$
are thresholds for opinions and beliefs, respectively.
They determine the tolerance of agents towards different opinions and beliefs.
\begin{example}
In the following examples, unless stated otherwise, we always consider that $\varepsilon=0.4$ and $\delta=3$.
In this case, only Bob is concordant with Alice as $d_B(B_\text{Alice}^0,B_\text{Bob}^0)=2\leq\delta$ and $d_O(O_\text{Alice}^0,O_\text{Bob}^0)=0.4\leq\varepsilon$.
On the opposite, $d_B(B_\text{Alice}^0,B_\text{Celia}^0)=2\leq\delta$ and $d_O(O_\text{Alice}^0,O_\text{Celia}^0)=0.6>\varepsilon$.
Hence, $C_\text{Alice}^0=\{\text{Bob}\}$.
\end{example}

\section{\uppercasing{Related work}{Related Work}}\label{sec:related-work}

We first discuss the definition of echo chambers and the way to measure them.
Opinion and belief propagation have been considered from the standpoint of opinion dynamics and belief change.
Thus, related models in these areas are introduced in this section as well as attempts to connect them.

\subsection{\uppercasing{Echo chambers in social networks}{Echo Chambers in Social Networks}}\label{sec:sota:ec}

Echo chambers in social networks describe a situation in which a group of persons strengthen their viewpoints by interacting only with persons who have opinions similar to theirs.
Precisely, `\emph{An echo chamber is [...] characterized by like-minded users predominantly interacting with each other. Within these echo chambers, users express and reinforce their beliefs on specific issues, thereby amplifying their shared viewpoints}'  \cite{Mahmoudi2024}.
We use the term `echo chamber' following the definition above and not as something inherently bad \cite{Coady_2024}.

Following this definition, echo chambers are groups of people satisfying three properties:
\begin{description}[topsep=0pt]
 \item[homogeneity:] group members share the same or very close opinions, or beliefs;
 \item[segregation:] group members mainly listen to people in the same group;
 \item[reinforcement:] group members' opinions or beliefs do not become more distant over time.
\end{description}

There are various ways to attempt at measuring echo chambers \cite{Hartmann_Wang_Pohlmann_Berendt_2025}:
social network analysis measures, e.g., centrality,
content-based metrics comparing nodes attributes, e.g., their beliefs or opinions,
homophily and polarization metrics comparing if nodes have common neighbors \cite{Interian_GMarzo_Mendoza_Ribeiro_2023},
or graph clustering and partitioning methods, also known as modularity measures \cite{Clauset_Newman_Moore_2004}.

Among the three properties above, homogeneity can be assessed by content-based metrics, segregation is rather complementary to homophily, but reinforcement is barely considered.
In particular, \cite{Sasahara2020} uses homogeneity and segregation, but not reinforcement.
To identify echo chambers according to this definition, we will propose a measure taking into account the three properties.

\subsection{\uppercasing{Opinion dynamics and echo chambers}{Opinion Dynamics and Echo Chambers}}\label{sec:sota:od}

How echo chambers arise has been explained using mathematical models such as opinion dynamics models.
Opinion dynamics studies how agents' opinions evolve through interactions~\cite{Dong2018}.
The basic idea of such models is that agents' opinions are affected by other agents' opinions.
\begin{example}
For example, if Alice has the utmost positive opinion toward the construction of tram lines,
when she listens to Bob and Celia's less positive opinions toward it, she may adopt a lower opinion than before.
\end{example}

One classical model that can be applied to it is the \emph{bounded confidence model}~\cite{Deffuant_Neau_Amblard_Weisbuch_2000,hegselmann2002-RAIODA}.
In \cite{Deffuant_Neau_Amblard_Weisbuch_2000}, two agents $a$ and $a'$ exchange their opinions asynchronously.
They update their opinions if their difference, i.e., $d_O(O_a^t,O_{a'}^t)$, is equal to or less than a threshold $\varepsilon\in\mathbb{R}$.
The updating rule can be described as:
\[
 O_a^{t+1} =O_a^t+\mu (O_{a'}^t-O_a^t), \text{ and }
 O_{a'}^{t+1} =O_{a'}^t+\mu (O_a^t-O_{a'}^t),
\]
\noindent such that $\mu$ is a constant between $0$ and $0.5$~\cite{Dong2018}.
\begin{example}
For instance, if Alice and Bob exchange their opinions, with $\mu=0.5$, $\varepsilon=0.4$ and the initial opinions previously mentioned, $O_\text{Alice}^1=1+0.5\times (0.6-1)=0.8$ and $O_\text{Bob}^1=0.6+0.5\times (1-0.6)=0.8$.
\end{example}
Based on this model, it is shown that low tolerance for dissimilar opinions, i.e., a small $\varepsilon$, prevents agents' opinions from converging.

Nowadays, opinion dynamics models based on bounded confidence have been proposed to explain how echo chambers arise~\cite{Sasahara2020,Cinus2022,Springsteen2024}.
A model with opinion threshold, timelines, and network rewiring has been proposed
to model online social networks, and particularly Facebook, realistically~\cite{Sasahara2020}.
In this model, each agent follows at least another agent.
At each iteration, one agent ($a$) is chosen randomly.
Then, agent $a$ updates its opinions and then can modify $N_a^t$ and submit new posts or reposts probabilistically.
The updating rule is a variation of the bounded confidence model involving all the agent's concordant neighbors:
\[
 O_a^{t+1}=
\begin{cases}
  \mu O_a^{t}+(1-\mu)\times\frac{\sum_{a'\in C_a^{t}} O_{a'}^{t}}{|C_a^t|} & \text{ if } C_a^t\neq\emptyset \\
  O_a^{t} & \text{ if } C_a^t=\emptyset
 \end{cases}
\]
\begin{example}
Assume that $\varepsilon=0.4$, $\delta=|\mathcal M(\top)|$ (so that beliefs do not matter) and $\mu=0.5$ and only Bob is concordant with Alice, then as before, $O_\text{Alice}^1$
 is $0.5\cdot 1.0+(1-0.5)\cdot 0.6=0.8$.
\end{example}

It is observed that the smaller the $\varepsilon$ threshold, the more echo chambers are created.
Moreover, it is reported that network rewiring, i.e., unfollowing and following, accelerates their appearance.

\fullversion{Using an extension of this model, Nagura and Akiyama~\cite{Nagura2025} show that the fewer topics are discussed by agents and the more important they are, the more echo chambers agents form.}{}

With the advance of large language models (LLMs), some models rely on LLM-based agents which explain reasons why they have such opinions~\cite{Ohagi2024,wang2024}.
However, how agents' beliefs are changed over iterations has not been analyzed.

\subsection{\uppercasing{Distributed belief change}{Distributed Belief Change}}\label{sec:sota:bc}

In the field of belief change \cite{ferme2018a}, many operators describe how to revise or merge beliefs.
\begin{example}
For example, assume that Alice believes that `if tram lines are built, then there will be less traffic jams ($p\rightarrow q$)' and Bob, believes that `if tram lines are built, then taxes will increase ($p\rightarrow \neg r$)'.
Then, after Alice listens to Bob, she may believe that the construction leads to less traffic jams \emph{and higher taxes} by taking Bob's beliefs into account.
\end{example}
However, little work considers belief propagation in a network.

Belief revision games (BRG)~\cite{Schwind_Inoue_Bourgne_Konieczny_Marquis_2015} define procedures, similar to those of opinion dynamics, for merging beliefs.
Such an operation, $R$, is defined as%
\footnote{$d^H$ is the Hamming distance between two interpretations, i.e., $d^H(m,m')=|\{p\in\mathcal{P};(m\models p\land m'\not\models p)\lor (m\not\models p\land m'\models p)\}|$.}:
\begin{equation}\label{eq:r3}
  R(\textbf{B})=\left[ \argmin_{m\in\mathcal{M}(\top)}\sum_{B\in\textbf{B}} 
\min_{m'\in\mathcal{M}(B)}d^H(m,m')
 \right]
\end{equation}
\noindent such that $\textbf{B}$ is a set of belief sets, those from the neighbors, to be merged and $\mathcal{M}(\top)$ is the set of all the interpretations on $\mathcal P$.
$R$ returns consistent beliefs \cite{Schwind_Inoue_Bourgne_Konieczny_Marquis_2015}.
\begin{example}
For example, Alice can alter her beliefs by revising her current beliefs by Bob's beliefs.
Hence, her new beliefs are $R(B_\text{Alice}^0,B_\text{Bob}^0)=\{p\to (q\land\lnot r)\}$ or $(p\land q\land\lnot r)\lor (\lnot p\land q\land r)\lor (\lnot p\land q\land\lnot r)\lor (\lnot p\land\lnot q\land r)\lor (\lnot p\land\lnot q\land\lnot r)$ in FDNF.
\end{example}

BeliefFlow~\cite{schwind2024a} is another framework which enables beliefs to be updated asynchronously.
In this framework, agents reach a consensus if the network they form is strongly connected.
None of these procedures allow agents to modify the network.
Network rewiring has been considered in the network knowledge base approach on a simpler belief description language \cite{Gallo2021}.
These work have not considered the impact of belief change on the formation of echo chambers, which are investigated here.

\subsection{\uppercasing{Models evolving opinions and beliefs}{Models Evolving Opinions and Beliefs}}\label{sec:sota:joinob}

In spite of socio-psychological evidence of their relation \cite{Oktar_Lombrozo_2025,wen2025a}, there has been few models explicitly connecting changing opinions and beliefs as defined in opinion dynamics and belief propagation.
Some models constrain Boolean public opinions with common integrity constraints expressed logically \cite{botan2019}.
Other work has used an influence matrix between opinions in order to enforce their dependencies \cite{friedkin2016a,nedic2019a}.
However, these constraints do not have the status of beliefs and do not change.

In \cite{tang2021a}, the distinction is made between beliefs and experience.
Experience is modeled as data incompletely sampling the environment.
From samples, agents learn beliefs, under the form of a classifier based on feature correlation.
Then they explicitly exchange both data and classifier with their neighbors.
They adopt neighbor classifiers only if it is more accurate on their own experience.
This procedure is called `epistemic rationality', meaning that agents try to maintain as much coherence as possible between their beliefs and their experience.
Although this work does not deal with opinions, only belief and data, it provides the notion of internal coherence of agents' beliefs and experience.

We have introduced a way to combine and synchronise opinions and beliefs \cite{kataoka2026a}.
This has been extended in \cite{kataoka2025a} by using confidence-bounded propagation and network rewiring in order to reproduce results of \cite{Sasahara2020}.
It shows that the connection between the two processes `reinforces' echo chambers based on global measures such as maximal distance between belief/opinion, number of `communities' (disregarding the network structure) and average number of discordant neighbors \cite{kataoka2025a}.
Although the second may be considered a measure of segregation, as in \cite{Sasahara2020}, reinforcement is completely left aside.
Moreover, this work does not introduce a precise definition of echo chambers.

Here, we refine this model and introduce a precise definition of echo chambers so that it is possible to count them and measure properties respective to these echo chambers instead of global properties.
This enables new observations on a comparable experiment and more precision in the results.
Moreover, we are able to contrast the results obtained when agents synchronize their beliefs and opinions with respect to when they do not do it. 

\section{\uppercasing{A model for opinion and belief propagation and integration}{A Model for Opinion and Belief Propagation and Integration}}\label{sec:model}

We propose a model in which opinions and beliefs are distinguished: opinions represent how positive or negative an agent considers a topic, while beliefs describe what it regards as true.

Agents change their opinions and beliefs with respect to four procedures ({\opiniondynamics}, {\opinionform}, {\beliefrevision}, and {\beliefalign}).
These functions map an agent's cognitive state to a new state.
We separate them in two types of functions:
belief and opinion propagation (Section~\ref{sec:propagation}) and
coherence restoration (Section~\ref{sec:coherence}).

\subsection{Propagation}\label{sec:propagation}

Agents can revise their opinions by taking other agents' opinions into account:
this corresponds to classical opinion dynamics models.
Similarly, beliefs can be revised based on others' beliefs:
this process is formalized by classical belief revision.
Hence, we simply use existing procedures.

\subsubsection{\uppercasing{Opinion dynamics}{Opinion Dynamics}}\label{sec:model:od}

In the opinion dynamics process (\opiniondynamics), agents update their current opinions
using the current opinions of their concordant neighbors.
Thus, $OD$ is defined as:
{\setlength{\tabcolsep}{1pt}
  \begin{align}
 OD(\langle O_a,B_a\rangle)=
 \begin{cases}
  \left\langle\mu O_a+(1-\mu)\frac{\sum_{a'\in C_a^t} O_{a'}}{|C_a^t|},B_a \right\rangle\!&\!\text{if } C_a^t\neq\emptyset \\
  \langle O_a,B_a \rangle\!&\!\text{if } C_a^t=\emptyset
 \end{cases}
 \label{eq:od}
\end{align}}
\noindent such that $\mu\in\interval{0}{1}$.
This procedure corresponds to the opinion updating process in \cite{Sasahara2020} which is based on the bounded confidence model (Section~\ref{sec:sota:od}).
\begin{example}
Hence, if Alice performs $OD$ as first operation, her cognitive state becomes $\langle 0.8,B_\text{Alice}^0\rangle$.
\end{example}
  
\subsubsection{\uppercasing{Belief revision}{Belief Revision}}\label{sec:model:br}

In the belief revision process (\beliefrevision), agents update their beliefs
using the current beliefs of their concordant neighbors.
Thus, $BR$ is defined as:
\begin{align}
 BR(\langle O_a,B_a\rangle)=\langle O_a, R(B_a,B_{a_1},\ldots,B_{a_n})\rangle
 \label{eq:br}
\end{align}
\noindent such that $C_a^t=\{a_1,\ldots,a_n\}$ are $a$'s concordant neighbors.
$R$ is the
operation defined in Section~\ref{sec:sota:bc}, chosen as the most regular operation among those of \cite{Schwind_Inoue_Bourgne_Konieczny_Marquis_2015}.
\begin{example}
Accordingly, if Alice updates her beliefs with concordant agents' beliefs (here Bob's beliefs) following $BR$, her state would become $\langle O_\text{Alice}^0, \{p\to (q\land\lnot r)\}\rangle$.
\end{example}

Setting $\delta=|\mathcal{M}(\top)|$ (resp.\ $\varepsilon=1$), in the definition of $C^t_a$, means that agents do not ignore their neighbors based on beliefs (resp.\ opinions), thus performing {\opiniondynamics} (resp.\ \beliefrevision) corresponds to the bounded confidence model \cite{Deffuant_Neau_Amblard_Weisbuch_2000} (resp.\ BRG \cite{Schwind_Inoue_Bourgne_Konieczny_Marquis_2015}).

\subsection{\uppercasing{Coherence between opinions and beliefs}{Coherence Between Opinions and Beliefs}}\label{sec:coherence}

After propagation, agents' opinions and beliefs may not correspond to each other.
\begin{example}
For example, after Alice listens to Bob's opinion, she may have a less positive opinion toward the construction of tram lines, but this did not affect her beliefs.
Conversely, she may have changed her beliefs from Bob's without changing her opinions.
Hence depending on whether opinions ($OD$) or beliefs ($BR$) have been propagated, she may found herself in a state, $\langle 0.8, B_\text{Alice}^0\rangle$ or $\langle O_\text{Alice}^0, \{p\rightarrow q\wedge \neg r\}\rangle$, which may need to be adjusted.
\end{example}
This loss of coherence may be uncomfortable for agents as they may not be able to justify their opinions from their beliefs.

In order to restore coherence, we use two synchronization operations: opinion formation from beliefs (\opinionform) and belief alignment with opinions (\beliefalign).
The former can be used to adapt opinions when beliefs have changed and the latter to adapt beliefs to new opinions.
Because opinions and beliefs are not direct translations of each other, these operations are based on
\emph{values} that agents may hold.

\subsubsection{\uppercasing{Cultural values}{Cultural Values}}\label{sec:values}

Cultural values help agents to determine whether something is `good' or `bad'~\cite{Schwartz1999,mercuur2019a}.
They are used here to determine preferences between different possible states of the world.
This is rendered as a map $V$ from interpretations in $\mathcal{M}(\top)$ to real numbers between $0$ and $1$.
Intuitively, the cultural values toward an interpretation $I$ is close to $0$ (resp.\ $1$), if the agent considers $I$ as bad (resp.\ good).
For example,
Table~\ref{tbl:values} represents values $V$ which consider worlds consistent with $p$, then $q$, then $r$, as good.
\begin{example}
In the example considered so far, these values mean that agents prefer to build tram lines more than reducing traffic jams and reducing traffic jams more than reducing taxes.
\end{example}

\begin{table}
 \centering
\caption{A score $V$ is provided to each Boolean assignment of three atoms.}\label{tbl:values}
\begin{tabular}{c|cccccccc}
 \toprule
 $p$ & t & t & t & t & f & f & f & f \\
 $q$ & t & t & f & f & t & t & f & f \\
 $r$ & t & f & t & f & t & f & t & f \\\midrule
 $V$ & $\frac{7}{7}$ & $\frac{6}{7}$ & $\frac{5}{7}$ & $\frac{4}{7}$ & $\frac{3}{7}$ & $\frac{2}{7}$ & $\frac{1}{7}$ & $\frac{0}{7}$  \\
\bottomrule
\end{tabular}
\end{table}

Such values allow agents to form their opinions from beliefs and vice versa as discussed below.

\subsubsection{\uppercasing{Opinion formation}{Opinion Formation}}

In the opinion formation process, an agent revises its opinions based on its beliefs.
\begin{example}
For example, Alice may want to lower her opinions towards trams as she has altered her beliefs after Bob's.
\end{example}
This consists first in generating an opinion from the current beliefs ($\phi$ is the topic on which to form an opinion):
\begin{align}
 v(B)=\frac{\sum_{m\in \mathcal{M}(R(B,\phi))}V(m)}{|\mathcal{M}(R(B,\phi))|}.\label{eq:v}
\end{align}
This means that an agent forms its opinion about the topic $\phi$ from its beliefs $B$
based on how valuable the models of the revision of $B$ by $\phi$ are.
The divisor of $v(B)$ is never $0$ since $R$ returns consistent beliefs \cite{Schwind_Inoue_Bourgne_Konieczny_Marquis_2015}.
\begin{example}
For example, after Alice has processed $BR$ with Bob, her cognitive state is $\langle 1.0, \{p \rightarrow (q \wedge \neg r)\}\rangle$ (see Section~\ref{sec:model:br}).
To adapt her opinions towards $\phi=p$ from her new beliefs,
she first revises her current beliefs by $\phi$:
$R(p \rightarrow (q \wedge \neg r), \phi)=p\land q\land\lnot r$.
Assume that her values $V$ are those defined in Table~\ref{tbl:values},
she can compute the mean of the values $V(m)$ where $m\in\mathcal{M}(R(p \rightarrow (q \wedge \neg r), \phi))$, hence $v(p \rightarrow (q \wedge \neg r))=\frac{6}{7}$.
\end{example}

The generated opinion is then aggregated to the agent's current opinion. 
The function $OF$ is thus defined as:
\begin{align}
OF(\langle O_a,B_a\rangle)=\langle\alpha O_a+(1-\alpha)\times v(B_a),B_a\rangle
 \label{eq:of}
\end{align}
with $\alpha\in\interval{0}{1}$ controlling the inertia to the changing process.
This is a simple way to balance inertia and willingness to change opinions.
\begin{example}\label{ex:of}
For example, if Alice had opinion $1.0$ toward $\phi$ and $\alpha=\frac{1}{2}$, i.e., her new opinion is affected by her current opinion and beliefs equally.
Then, she could obtain the new opinion as:
$ 0.5\cdot 1.0 + (1-0.5) \cdot \frac{6}{7} = \frac{13}{14} \approx 0.93.$
Hence, her new cognitive state is $\langle\frac{13}{14},\{p \rightarrow (q \wedge \neg r)\}\rangle$.
This result means that she considers the topic less positively than before.
\end{example}

\subsubsection{\uppercasing{Belief alignment}{Belief Alignment}}

In the belief alignment process, agents revise their beliefs based on their opinions.
\begin{example}
For example, after Alice has changed her opinion as influenced by her neighbors, she may want to alter her beliefs to align with her opinions.
\end{example}
This is achieved by filtering beliefs in two steps.
First, agents select the beliefs which minimize the distance between their current opinions and the opinions based on beliefs from the function $v$ of Equation~\ref{eq:v}.
Second, agents adopt those consistent beliefs which minimize the distance between their current beliefs and those beliefs selected in the previous step.
This assumes that agents only adopt consistent beliefs.

Thus, the function $BA$ is defined as:
\begin{align}
  BA(\langle O_a,B_a\rangle) & =\langle O_a,\argmin_{B\in\mathcal{L_P}(O_a)}d_B(B,B_a)\rangle
 \label{eq:ba}\\
\text{such that~~~~~~~~~}\qquad
 \mathcal{L_P}(O)&=\argmin_{B\in\mathrm{FDNF}^*(\mathcal{L_P})}d_O(O,v(B)) \notag
\end{align}
\noindent and $\mathrm{FDNF}^*(\mathcal{L_P})$ is the set of consistent FDNF formulas (excluding the empty disjunction).
If
several beliefs
minimize $d_B(B,B_a)$ in Equation~\ref{eq:ba}, the aligned result is chosen randomly among them.

\begin{example}
For instance, after having processed $OD$ (Section~\ref{sec:model:od}), Alice's state is $\langle 0.8, \{p\to q\}\rangle$.
To align her beliefs to her new opinions, Alice first considers which beliefs are the best to support her opinions, i.e., which beliefs minimize the distance between her current opinion and the opinion which would be formed from her beliefs ($\mathcal {L_P}(0.8)$).
\fullversion{Here, the set of the beliefs supporting her opinion is
$\mathcal {L_P}(0.8)=\{B\in\mathcal{L_P};\exists B'\in\mathcal{L_P},(p\land q\land r)\lor (p\land q\land \lnot r)\lor (p\land \lnot q\land\lnot r)\lor (\lnot p\land B')\}$.}{}
Then, she picks one of the closest beliefs $B\in\mathcal{L_P}(0.8)$ to her current beliefs $B_\text{Alice}^0$ as possible.
In this case, $B=\{p\wedge r\rightarrow q\}$ (in FDNF: $(p\land q\land r)\lor (p\land q\land \lnot r)\lor (p\land \lnot q\land\lnot r)\lor (\lnot p\land q\land r)\lor (\lnot p\land q\land\lnot r)\lor (\lnot p\land\lnot q\land r)\lor (\lnot p\land \lnot q\land\lnot r)$) minimizes the distance to $B_\text{Alice}^0$, as $d_B(B,B_\text{Alice}^0)=1$.
Hence, when Alice performs belief alignment, her new cognitive state is $\langle 0.8,B\rangle$.
\end{example}

\subsection{Protocol}\label{sec:interactions}

Agents behave following a procedure, adapted from \cite{Sasahara2020}, combining these four operations.
At time $t$, the set $A^t\subseteq A$ of activated agents is determined by drawing agents in $A$ with probability $p_\text{activate}$.
Each non active agent ($a\notin A^t$) does nothing at time $t$;
active agents update their opinions and beliefs, and possibly their
neighbors.

We distinguish four experimental processes to test our hypotheses: (\ref{eq:odonly}) (resp. (\ref{eq:bronly})) only performs opinion dynamics (resp. belief revision); (\ref{eq:both}) performs both operations and (\ref{eq:coherence}) applies, in addition, opinion formation and belief alignment to synchronize them:
\begin{align}
 \langle O_a^{t+1},B_a^{t+1}\rangle &=OD(~\langle O_a^t,B_a^t\rangle~), \tag{odonly}\label{eq:odonly}\\
 \langle O_a^{t+1},B_a^{t+1}\rangle &=BR(~\langle O_a^t,B_a^t\rangle~), \tag{bronly}\label{eq:bronly}\\
 \langle O_a^{t+1},B_a^{t+1}\rangle &=BR(~OD(~\langle O_a^t,B_a^t\rangle~)~), \tag{both}\label{eq:both}\\
 \langle O_a^{t+1},B_a^{t+1}\rangle &=OF(~BR(~BA(~OD(~\langle O_a^t,B_a^t\rangle~)~)~)~). \tag{coherence}\label{eq:coherence}
\end{align}
\hiroForLater{Perform $OF$ and $BA$ until opinions and beliefs reach stability?}
All functions are executed synchronously among
active agents.

Then, as in \cite{Sasahara2020}, with the probability $p_\text{rewire}$, agent $a\in A^t$ tries to modify its set of neighbors.
Intuitively, this is achieved by removing from $N_a^t$ one agent $a'$ that is not concordant and adding one agent $a''$ that is not a neighbor or $a$ itself to it.
More formally, $a$ considers the sets $F_a^t\subseteq A$ and $U_a^t\subseteq A$ of agents that can be followed and unfollowed at time $t$, respectively.
They are defined as:
\[
 U_a^t = N_a^t\setminus C_a^t  \text{ and }
 F_a^t = \{a'\in A; a\neq a'\land a'\notin N_a^t\}.
\]
If the two sets are non empty, $a$ can choose the unfollowed agent $a'\in U_a^t$ and the followed agent $a''\in F_a^t$ randomly, and update its set of neighbors as follows:
\begin{align*}
 N_a^{t+1}=(N_a^t\setminus\{a'\})\cup\{a''\}.
\end{align*}
Otherwise, the set of neighbors is not updated, i.e., $N_a^{t+1}=N_a^t$.
This entails that for each $a\in A$, the size of $N_a^t$ remains constant over $t$.
\begin{example}
For example, when Alice performs the rewiring, the unfollowed agent $a'$ is Celia because $U_\text{Alice}^0=\{\text{Celia}\}$.
Moreover, in the network presented in Figure~\ref{fig:intuitive-scenario-network} (left), David is the only agent who is not a neighbor of Alice, i.e., $F_\text{Alice}^0=\{\text{David}\}$.
Hence, Alice can update her neighbors as:
\[
 N_\text{Alice}^1=(N_\text{Alice}^0\setminus\{\text{Celia}\})\cup\{\text{David}\}=\{\text{Bob},\text{David}\}.
\]
Figure~\ref{fig:intuitive-scenario-network} (right) shows the network after the rewiring.
\end{example}

\begin{table}\setlength{\tabcolsep}{3pt}
 \centering
 \caption{Configuration of the parameters.
$\measure{eo}^t_{X}$ and $\measure{eb}^t_{X}$ mean the measure $\measure{eo}^t$ and $\measure{eb}^t$ obtained in the setting $X$, respectively.}
 \label{tbl:parameters}
 \begin{tabular}{c|cccc}
  \toprule
  Hypothesis & H0 & H1 & H2 & H3 \\
\midrule
  Setting & S0 & S1 & S2 & S3 \\
\midrule
  protocol & (\ref{eq:odonly}) & (\ref{eq:bronly}) & (\ref{eq:both}) & (\ref{eq:coherence}) \\
  $\varepsilon$ & $\{0.05,0.1,\ldots,0.5\}$ & $1$ & $\{0.05,0.1,\ldots,0.5\}$ & $\{0.05,0.1,\ldots,0.5\}$ \\
  $\delta$ & $8$ & $\{1,2,\ldots,7\}$ & $\{1,2,\ldots,7\}$ & $\{1,2,\ldots,7\}$ \\
\midrule
  \#Sims  & $20\cdot 10=200$ & $20\cdot 7=140$ & $20\cdot 10\cdot 7=1400$ & $20\cdot 10\cdot 7=1400$  \\
\midrule
  \multirow{2}{*}{Test}  & \multirow{2}{*}{$\measure{eo}^T_{S0}>0$} & \multirow{2}{*}{$\measure{eb}^T_{S1}>0$} & $\measure{eo}^T_{S2}\geq\measure{eo}^T_{S0}$ &  $\measure{eo}^T_{S3}\geq\measure{eo}^T_{S2}$ \\
  &  &  & and $\measure{eb}^T_{S2}\geq\measure{eb}^T_{S1}$ & and $\measure{eb}^T_{S3}\geq\measure{eb}^T_{S2}$ \\
  \bottomrule
 \end{tabular}
\end{table}

\section{Measures}\label{sec:measures}

In \cite{Sasahara2020}, the emergence of echo chambers is shown
visually
using plots of the evolution of message diversity, agents' opinions, and the network of agents over time.
However, these plots are generated from results of one single experiment.

In the same paper, the effects of $\varepsilon$ on the number of final opinion peaks and final opinion distance are presented based on 20 experiments which share the same parameters.
These measures show how opinions are polarized and the network segregated.
Although polarization may be
a measure of homogeneity, reinforcement is ignored, so this should not allow us to conclude on echo chambers.
Results have been confirmed in \cite{kataoka2025a} but still with measures not explicitly identifying echo chambers.

The way to identify echo chambers is to partition the network into groups and to count those satisfying the properties of homogeneity, segregation and reinforcement (see Section~\ref{sec:sota:ec}).
Of course, it is not practicable to test all partitions,
thus we propose to use strongly connected components to identify echo chambers.
Basing measurements on strongly connected components make sense because:
(1) echo chambers are expected to be connected,
(2) they are maximal, and
(3) reinforcement is usually obtained through feedback, which needs strong connection.
Nevertheless, strongly connected components may be replaced by other structures, partitions or not.
The effect of this choice is left for further investigation.
Hereafter, \emph{component} denotes strongly connected components and $\sccs$ is the set of components at time $t$.

During the experiments, the following measures are recorded.

\paragraph{Segregation}
The \emph{segregation} of a component is measured by the ratio of the edges whose sources are in a component $C$ and whose targets
are not in $C$ to the edges whose sources are in $C$:
\[
 L^t(C)=\frac{|\{\langle a,a'\rangle\in N^t;a\in C\land a'\notin C\}|}{|\{\langle a,a'\rangle\in N^t;a\in C\}|}.
\]
Hence, the lower this measure, the more segregated the network.

\paragraph{Homogeneity}
To assess how \emph{homogeneous} agents in a component $C\in \sccs$ are, we measure the maximal distance between opinions (resp.\ beliefs) of agents which belong to the component:
\begin{align*}
  M_O^t(C) &= \max_{a,a'\in C} d_O(O_a^t,O_{a'}^t),\\
  M_B^t(C) &= \max_{a,a'\in C} d_B(B_a^t,B_{a'}^t).
\end{align*}
The lower these measures, the more homogeneous the component.

\paragraph{Reinforcement}
To assess if a component $C$ satisfies \emph{reinforcement}, we test whether its homogeneity measure (maximal distance between agents' opinions or beliefs) decreases during its lifespan.  
Let $\interval{t_C}{t}$ be the maximal window during which a component $C\in \sccs[t]$ exists,
then, the measure is defined as:\\[-10pt]
\begin{align*}
  D_O^t(C) &\equiv \forall s\in \interval[open right]{t_C}{t},M_O^s(C)\geq M_O^{s+1}(C)\\
  D_B^t(C) &\equiv \forall s\in \interval[open right]{t_C}{t},M_B^s(C)\geq M_B^{s+1}(C)
\end{align*}

\paragraph{Echo chambers}

Based on the above measures, we can compute the number of echo chambers at time $t$ as those components that are segregated and whose opinions and beliefs are homogeneous and reinforcing:
\begin{align*}
 \measure{eo}^t&=|\{C\in \sccs;\overbrace{L^t(C)\leq\theta}^\text{segregation}~\land~ \overbrace{M_O^t(C)\leq 10^{-5}}^\text{homogeneity}~\land~ \overbrace{D_O^t(C)}^{\mathclap{\text{reinforcement}}}\}|,\\
 \measure{eb}^t&=|\{C\in \sccs;L^t(C)\leq\theta~\land\quad M_B^t(C)=0 \ \  ~\land~ D_B^t(C)\}|.
\end{align*}
\noindent such that $\theta\in\interval[open]{0}{1}$.
The measures are made independent of $\varepsilon$ and $\delta$, which vary in experiments, by choosing stricter thresholds ($10^{-5}$ and $0$) for the homogeneity property.
In the following, we use $\theta=0.5$.
Here, $L^t(C)\leq 0.5$ indicates that at most half of the edges whose sources are in $C$, have their targets out of $C$.
This automatically disqualifies singletons (components $C$ such that $|C|=1$) as echo chambers.
In complement, we compute the proportion of agents involved in echo chambers.

\begin{example}
In Figure~\ref{fig:intuitive-scenario-network} (left), there are three strongly connected components, but only Celia who does not listen to any one could be an echo chamber, as Alice, Bob and David have significantly different opinions and beliefs.
After propagation and rewiring (right), Alice, Bob and David form a strongly connected component that is an echo chamber.
\end{example}

\section{Hypotheses}\label{sec:hypotheses}

We test the four following hypotheses taking into account the more elaborate definition of echo chambers.

First, the proposed model is defined based on that of \cite{Sasahara2020}.
So, if only opinions are updated, echo chambers should be observed:
\begin{description}[topsep=0pt,leftmargin=!,labelwidth=\widthof{\bfseries H0}]
\item[\hypothesis{0}] If opinions evolve independently from beliefs, 
  opinion echo chambers should arise.
\end{description}
Similarly with respect to \cite{kataoka2025a}, the same should occur for beliefs:
\begin{description}[topsep=0pt,leftmargin=!,labelwidth=\widthof{\bfseries H0}]
\item[\hypothesis{1}] If beliefs evolve independently from opinions, 
   belief echo chambers should arise.
\end{description}

Next, it is expected that independent opinion and belief propagation reinforce each other, through the double constraint put on concordant agents, and thus lead to more echo chambers:
\begin{description}[topsep=0pt,leftmargin=!,labelwidth=\widthof{\bfseries H0}]
\item[\hypothesis{2}] Connecting opinions and beliefs increases the number of echo chambers.
\end{description}
Finally, keeping opinions and beliefs coherent is expected to tighten the links between them and to induce more echo chambers than without keeping the coherence:
\begin{description}[topsep=0pt,leftmargin=!,labelwidth=\widthof{\bfseries H0}]
\item[\hypothesis{3}] Connecting opinions and beliefs and keeping them coherent increases the number of
echo chambers.
\hiroForLater{And further hypotheses if we see the effect of changing each parameters? (of course it is likely that this is for my thesis, not ECAI ...)}
\end{description}

\begin{table*}\setlength{\tabcolsep}{1pt}
 \centering
 \caption{Overall results for the four experimental settings and the initial setting. In parentheses for components are the percentages of agents belonging to such components.
Raw results are aggregated by computing the mean.}
 \label{tbl:average-ec}
 \begin{tabular}{c|c|cccc}
  \toprule
  Setting & initial &  S0 & S1 & S2 & S3 \\
   \midrule
  stabilized experiments & -- & $200$ {\footnotesize ($100\%$)} & $140$ {\footnotesize ($100\%$)} & $1389$ {\footnotesize ($99.2\%$)} & $1390$ {\footnotesize ($99.3\%$)} \\
   \midrule
  components ($|\mathcal{S}^T|$)           & $3.35$ {\footnotesize ($100\%$)}  & $4.29$ {\footnotesize ($100\%$)}  & $7.48$ {\footnotesize ($100\%$)}  & $7.97$ {\footnotesize ($100\%$)}  & $8.29$ {\footnotesize ($100\%$)} \\
  singletons                               & $2.35$ {\footnotesize ($2.35\%$)} & $2.35$ {\footnotesize ($2.35\%$)} & $4.39$ {\footnotesize ($4.39\%$)} & $4.25$ {\footnotesize ($4.25\%$)} & $4.29$ {\footnotesize ($4.29\%$)} \\
\fullversion{
  segregated components ($L^T$)            & $1.00$ {\footnotesize ($97.7\%$)} & $1.71$ {\footnotesize ($97.2\%$)} & $2.29$ {\footnotesize ($93.7\%$)} & $2.87$ {\footnotesize ($93.7\%$)} & $3.30$ {\footnotesize ($94.0\%$)} \\
  opinion homogeneous ($M^T_O$) components & $2.35$ {\footnotesize ($2.35\%$)} & $4.29$ {\footnotesize ($100\%$)}  & --                & $7.76$ {\footnotesize ($92.8\%$)} & $7.94$ {\footnotesize ($87.4\%$)} \\
  belief homogeneous ($M^T_B$) components  & $2.35$ {\footnotesize ($2.35\%$)} & --                & $7.19$ {\footnotesize ($85.4\%$)} & $7.58$ {\footnotesize ($80.9\%$)} & $7.96$ {\footnotesize ($86.1\%$)} \\
  opinion reinforcing ($D^T_O$) components & --                & $4.29$ {\footnotesize ($100\%$)}  & --                & $7.97$ {\footnotesize ($100\%$)}  & $8.29$ {\footnotesize ($100\%$)} \\
  beliefs reinforcing ($D^T_B$) components & --                & --                & $7.48$ {\footnotesize ($100\%$)}  & $7.97$ {\footnotesize ($100\%$)}  & $8.29$ {\footnotesize ($100\%$)} \\ \midrule }{}
  opinion echo chambers ($\measure{eo}^T$) & $0.00$ {\footnotesize ($0\%$)}    & $1.71$ {\footnotesize ($97.2\%$)} & --                & \textcolor{blue}{$2.59$ {\footnotesize ($85.7\%$)}} & \textcolor{blue}{$1.97$ {\footnotesize ($19.3\%$)}} \\
  beliefs echo chambers ($\measure{eb}^T$) & $0.00$ {\footnotesize ($0\%$)}    & --                & \textcolor{blue}{$1.91$ {\footnotesize ($70.9\%$)}} & \textcolor{blue}{$2.41$ {\footnotesize ($67.0\%$)}} & \textcolor{blue}{$2.85$ {\footnotesize ($69.5\%$)}} \\
   \midrule
  Test & -- & $\surd$ & $\surd$ & $\surd$ and $\surd$ &  \textcolor{blue}{$\times$} and $\surd$ \\
\bottomrule
 \end{tabular}
\end{table*}

\section{\uppercasing{Experimental settings}{Experimental Settings}}\label{sec:experimental-settings}

In order to test each of the four hypotheses of Section~\ref{sec:hypotheses}, we perform four experiments under specific
conditions described hereafter.

\hiroForLater{Maybe not for ECAI but my thesis: we have introduced many parameters and thus we will have to see whether changing them affects the results.
But full-factorial plan over all pairs of parameters can yield too many experiments.
Thus, I think we will compare the basic set of experiments (this can be based on experiments in Sasahara) and other sets which share the same parameters except for one.}

We fix $\mathcal P$ to $\{p,q,r\}$ and the topic $\phi$ to $p$.
Moreover, we fix $|A|=100$ and $|N^0|=400$ to follow the same conditions
as those of \cite{Sasahara2020,kataoka2025a}.
We always use $T=5000$ iterations.

In addition, $p_\text{active}=p_\text{rewire}=0.5$ and $\mu=0.5$ to perform experiments under the same conditions as the experiments used to generate Figure~3 in \cite{Sasahara2020}.
We fix $\alpha=0.5$ so that the effect of an agent's beliefs on its opinions is the same as the effect of other agents' opinions on it.
Finally, the cultural values $V$ presented in Table~\ref{tbl:values} is shared among all of the agents and does not change over time.

The initial state is generated by drawing randomly $|N^0|$ edges among $|A|$ agents, so that each agent has at least one neighbor.
Agents are assigned random beliefs and opinions.
We use 20 different seeds to control the randomness of both the initial network and the propagation process.
Before starting interacting with each other, all agents perform $BA\circ OF$ until their opinions and beliefs become stable, i.e., each agent's beliefs do not change and the difference between opinions before and after the operation is less than or equal to $10^{-5}$.
Thus, agents are in a coherent state at the beginning.

$\delta$ is taken from the set $\{1,2,\ldots,|\mathcal M(\top)|\}=\{1,2,\ldots,8\}$.
$\varepsilon$ is taken from the set $\{0.05,0.1,\ldots,0.5\}$ so that the experiments correspond to those performed in \cite{Sasahara2020} and \cite{kataoka2025a}.

Table~\ref{tbl:parameters} relates
controlled parameters
to hypotheses.
The different experiment sizes (\#Sims) are the consequence of the controlled variables with respect to the hypotheses: when a parameter is frozen, all values lead to the same result.
\hiroForLater{Probably not for ECAI: to test how other parameters work, modify one parameter from the experiments for H2 and compare the results.}

\section{\uppercasing{Results and discussion}{Results and Discussion}}\label{sec:results}

Results of the four experiments are summarized in Table~\ref{tbl:average-ec}.
Thereby, `stabilized experiments' is the number of experiments in which agents' opinions and beliefs did not change during the last 500 iterations, i.e., $B_a^t\equiv B_a^T$ and $|O_a^t-O_a^T|\leq 10^{-5}$.
Although agents' cognitive states did not change, the network itself may change due to
network rewiring,
albeit without affecting agent states.

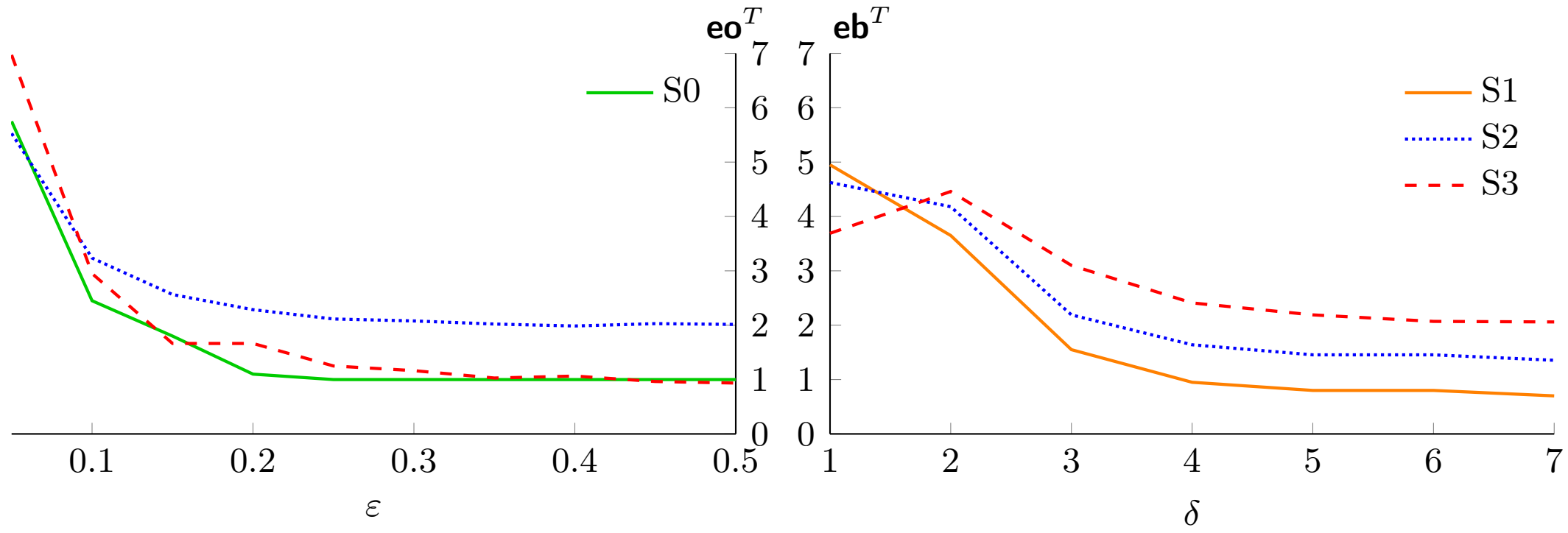
\begin{figure}[h!]
  \centering
 \begin{tikzpicture}[font=\footnotesize]
   \pgfplotsset{legend style={draw=none}}
      \begin{axis}[
        width=.55\linewidth, 
        height=5cm,
          xlabel=$\varepsilon$,
          xmin=0.05,
          xmax=0.5,
          axis x line*=bottom,
          xtick distance=.1,
          ylabel=$\measure{eo}^T$, 
          ymin=0,
          ymax=7,
          axis y line*=right,
          ytick distance=1,
          ylabel style={at={(current axis.north east)},anchor=south,rotate=270},
          major tick length=1mm
          ]
        \addplot[color=green!80!black, thick] plot table[x=eps,y=odonly,col sep=comma] {data/20260217-BROD-a-eo.csv}; 
        \addplot[color=blue,densely dotted, thick] plot table[x=eps,y=both,col sep=comma] {data/20260217-BROD-a-eo.csv}; 
        \addplot[color=red,dashed, thick] plot table[x=eps,y=coherence,col sep=comma] {data/20260217-BROD-a-eo.csv}; 
        \legend{S0}
      \end{axis}

      \begin{scope}[xshift=.5\linewidth]
      \begin{axis}[
        width=.55\linewidth, 
        height=5cm,
          xlabel=$\delta$,
          xmin=1,
          xmax=7,  
          axis x line*=bottom,
          xtick distance=1,
          ylabel=$\measure{eb}^T$,
          ymin=0,
          ymax=7,
          axis y line*=left,
          ytick distance=1,
          ylabel style={at={($(current axis.north west)+(-.48\linewidth,0)$)},anchor=south,rotate=270},
          major tick length=1mm
        ]
        \addplot[thick,color=orange,legend=bronly] plot table[x=delta,y=bronly,col sep=comma] {data/20260217-BROD-a-eb.csv}; 
        \addplot[color=blue,densely dotted, thick] plot table[x=delta,y=both,col sep=comma] {data/20260217-BROD-a-eb.csv}; 
        \addplot[color=red,dashed, thick] plot table[x=delta,y=coherence,col sep=comma] {data/20260217-BROD-a-eb.csv}; 
        \legend{S1,S2,S3}
      \end{axis}
      \end{scope}
    \end{tikzpicture}
    \caption{Effect of $\varepsilon$, resp. $\delta$, on the number of opinion, resp. belief, echo chambers.
      More tolerant agents lead to less echo chambers, but connecting opinions and beliefs leads to more echo chambers.
    {\color{blue}\protect\hypersetup{allcolors=blue}The curve of opinion in S3 condition changed, w.r.t. \cite{kataoka2026c}.\protect\hypersetup{allcolors=black}}}
 \label{plot:result:eps-delta-echochambers}
\end{figure}

\smallskip\noindent\textbf{Hypothesis H0.}
In each simulation with S0, $\measure{eo}^T>0$, i.e., at least one echo chamber is created.
Thus, the results support Hypothesis H0.
In addition, all components (not only echo chambers), \textcolor{blue}{but one}, satisfy reinforcement.
Figure~\ref{plot:result:eps-delta-echochambers} (left plot, solid green curve) shows the decrease of echo chambers with the increase of tolerance (higher $\varepsilon$).
Hence, although we applied a stronger measure, results confirm those of \cite{Sasahara2020}.

\begin{figure*}[b!]
  \newcounter{subfigcounter}
  \setcounter{subfigcounter}{0}
\makeatletter
\renewcommand*{\thesubfigcounter}{\alph{subfigcounter}}
\makeatother
\centering
\begin{minipage}{0.475\textwidth}
\centering
 \begin{tikzpicture}
  \node at (0,0) {\scalebox{0.25}{\input{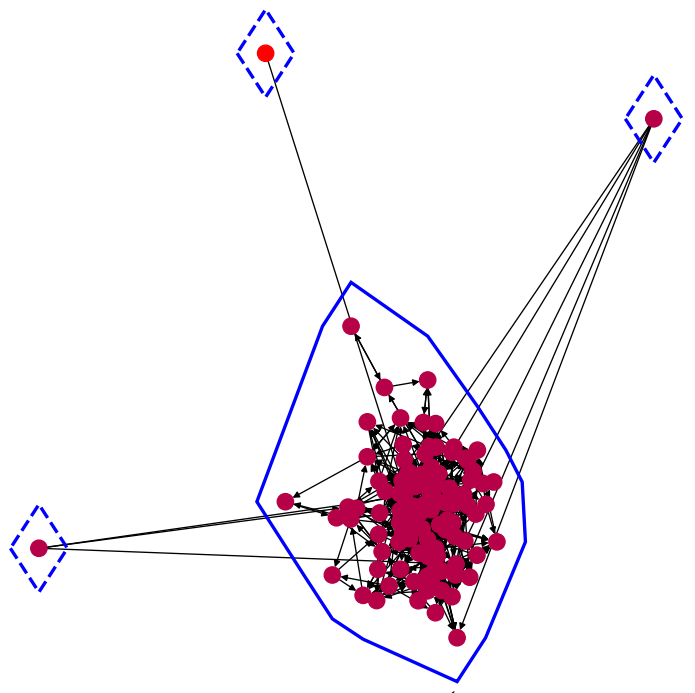}}};
 \end{tikzpicture}\\ [-5mm]
 {\small\xusecounter{subfigcounter}{it:networks:s0} Final network with S0 (Same color means same opinion).}
\end{minipage}
\hfill
\begin{minipage}{0.475\textwidth}
\centering
 \begin{tikzpicture}
  \node at (0,0) {\scalebox{0.25}{\input{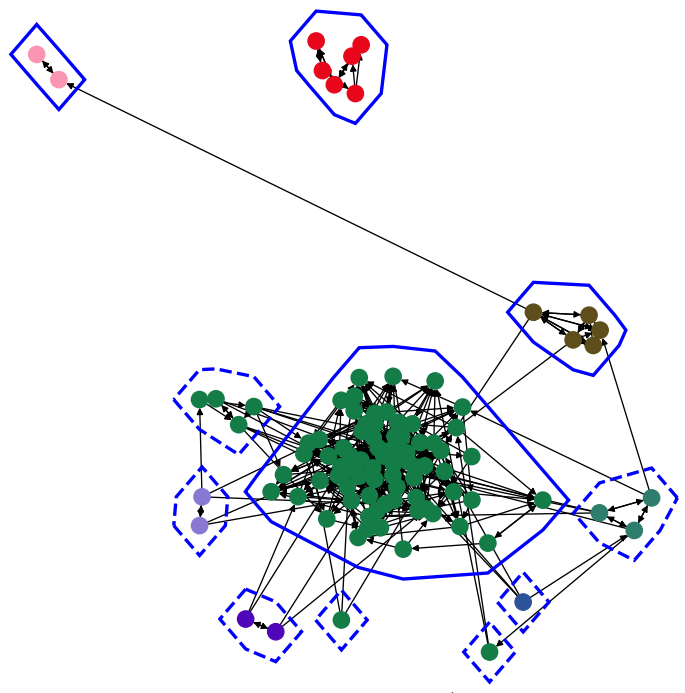}}};
 \end{tikzpicture}\\ [-5mm]
 {\small\xusecounter{subfigcounter}{it:networks:s1} Final network with S1 (Same color means same belief).}
\end{minipage}\\[-5pt]
\begin{minipage}{0.475\textwidth}
\centering
 \begin{tikzpicture}
  \node at (0,0) {\scalebox{0.25}{\input{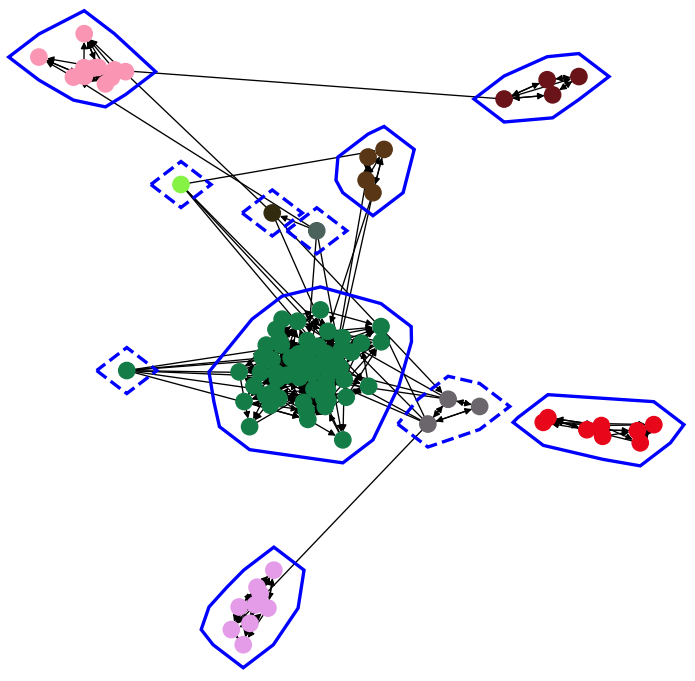}}};
 \end{tikzpicture}\\ [-5mm]
 {\small\xusecounter{subfigcounter}{it:networks:s2} Final network with S2 (Same color means same belief and opinion).}
\end{minipage}
\hfill
\begin{minipage}{0.475\textwidth}
 \centering
 \begin{tikzpicture}
  \node at (0,0) {\scalebox{0.25}{\input{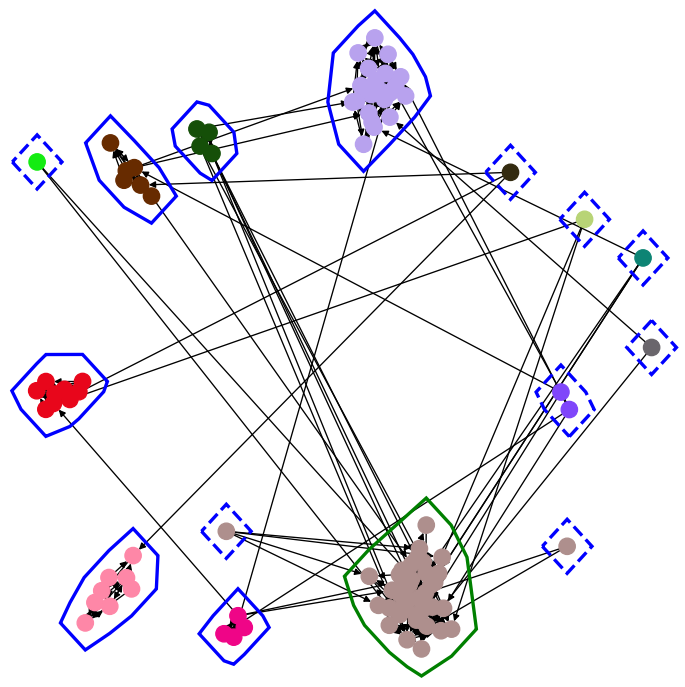}}};
 \end{tikzpicture}\\ [-5mm]
 {\small\xusecounter{subfigcounter}{it:networks:s3} Final network with S3 (Same color means same belief and opinion).
   \textcolor{blue}{The component circled in green is a belief echo chamber but not an opinion echo chamber.}}
\end{minipage}
\caption{Networks at $T$ from experiments using $\varepsilon=0.2$, $\delta=2$, and the same seed.}\label{fig:networks}
\end{figure*}

\xfigref{fig:networks}{it:networks:s0} shows what happens on a specific case: a
single
large component gathers
most agents with the same opinion.
A few singleton components listen to the echo chamber.
The echo chamber effect is huge in this case, as in many, as all agents have the same opinion.

\smallskip\noindent\textbf{Hypothesis H1.}
In each simulation with S1, but \textcolor{blue}{20}
out of 140 (\textcolor{blue}{86}\%), $\measure{eb}^T>0$.
These results show that we can observe belief echo chambers in most of the cases.
Thus, Hypothesis H1 is supported.

As for opinions, Figure~\ref{plot:result:eps-delta-echochambers} (right plot, solid orange curve) shows the number of echo chambers decrease with the increase of tolerance (higher $\delta$).

\xfigref{fig:networks}{it:networks:s1} shows the network scattered into many smaller components, with more echo chambers and more diverse beliefs.

\smallskip\noindent\textbf{Hypothesis H2.}
Table~\ref{tbl:average-ec} shows that, on average, connecting opinions and beliefs (S2) increases the number of opinion (S0) and belief (S1) echo chambers.
This supports Hypothesis H2.
In counterpart, the number of agents involved in these echo chambers, especially with respect to opinions, has decreased.
\xfigref{fig:networks}{it:networks:s2} shows that this is not only due to singleton components.

An interesting remark is that, the number of echo chambers increases without directly synchronizing opinions and beliefs.
This is explained by the rewiring of the network, which generates segregation.
Segregation, because agents start with coherent beliefs and opinions, tends to perpetuate coherence.
In consequence, echo chambers on both opinions and beliefs are closely related.

The echo chambers in \xfigref{fig:networks}{it:networks:s2} (and \ref{it:networks:s3}) are indeed homogeneous in both opinions and beliefs.

\begin{table}
  \setlength{\tabcolsep}{3pt}
 \centering
 \caption{Distribution of experiments with respect to the effect of maintaining coherence (S3) compared to not maintaining it (S2) on $\measure{eb}^T$ and $\measure{eo}^T$ (all other parameters being the same).}
 \label{tbl:cmp-eo-eb}
\begin{tabular}{c|ccc}
\toprule
\diagbox{$\measure{eo}^T$}{$\measure{eb}^T$}  & decreased & did not change & increased \\ \midrule
decreased & \textcolor{blue}{363 (25.9\%)} & \textcolor{blue}{311 (22.2\%)} & \textcolor{blue}{224 (16.0\%)} \\
did not change & \textcolor{blue}{0} & \textcolor{blue}{64 (4.6\%)} & \textcolor{blue}{184 (13.1\%)} \\
increased & 0 & \textcolor{blue}{2 (0.14\%)} & \textcolor{blue}{252 (18.0\%)} \\
\bottomrule
\end{tabular}

\end{table}

\smallskip\noindent\textbf{Hypothesis H3.}
\hiroForLater{I am planning to perform $OF$ and $BA$ until they stabilize (and do more interactions if necessary). NIY}
Table~\ref{tbl:average-ec} shows that simulations with S3 lead, on average, to more \textcolor{blue}{belief} echo chambers than those with S2 involving roughly the same proportion of agents.
\textcolor{blue}{For opinion echo chambers, S3 produced less and smaller echo chambers than S2.}
This would support Hypothesis H3 \textcolor{blue}{for belief echo chambers only; the hypothesis for opinion echo chambers is not supported}.
There is no direct relation between S2 and S3: pairwise comparison does not indicate a systematic increase \textcolor{blue}{or decrease}, as shown in Table~\ref{tbl:cmp-eo-eb}.
{\color{blue} It even shows that it makes belief echo chamber increase (660+377 out of 1400) and opinion echo chamber decrease (898+248 out of 1400).
Meanwhile, the average number of opinion echo chambers after synchronization ($1.97$) is still above the initial number of opinion echo chambers ($1.71$).}

\xfigref{fig:networks}{it:networks:s2} and (\ref{it:networks:s3}) show an increase of echo chambers (6 to 7).

\section{Conclusion}\label{sec:conclusion}

In social networks, symbolic beliefs are propagated as well as opinions.
To test their influence on echo chambers, we use a model in which agents exchange and update their opinions and/or beliefs using those of their neighbors.
Moreover, they maintain the coherence between their opinions and beliefs using
cultural
values.

We identify echo chambers based on components satisfying segregation, homogeneity and reinforcement of opinions or beliefs.
Experimental results confirm those obtained with different measures: the more tolerant agents are, the less echo chambers they create.
This also shows that joining opinions and beliefs generates more echo chambers.
This occurs even when agents do not explicitly maintain the coherence between their opinions and beliefs.

Hence, \textcolor{blue}{synchronizing} beliefs to opinions indeed affects both types of echo chambers. 
This does not prevent echo chambers from arising.
On the contrary, there are more \textcolor{blue}{belief} echo chambers, albeit smaller ones (see Figure~\ref{fig:networks}), involving slightly fewer agents.
Thus, it is worth asking if this strengthens or weakens echo chambers.
This could be tested by evaluating how the resulting echo chambers persist in the case of perturbations, i.e. locally modifying the graph topology or agent states.
\hiroForLater{
  That would be great to do it and first start by listing the kind of perturbations that can be applied:
  - adding edge, -suppresing edge, - adding edge from one new agent (we could also test agent introduction by neighbors in rewiring...).
}

\hiroForLater{
We did not consider the influence of different values and topics, which may lead to differentiated results.
Investigating such effects is left for future work.
}

\noindent\textbf{Data availability.} All experiments were performed with the \emph{SOBA} simulator \cite{soba}.
The experiment notebook is openly available from \textcolor{blue}{\cite{20260217-BROD-a}}.
It contains reference to software version, processing instructions, results, and analyses, including statistical tests.

\noindent\textbf{Ethical statement.}
The authors declare no competing interests.
The authors are not aware of any ethical issue with this work.

\subsection*{Acknowledgements}

This work was partially supported by the Fukuro INRIA-University of Tsukuba associate team and the Echo action of the NTU-UGA-UT trilateral center.
HK has been partially supported by the École Française de l'Intelligence Artificielle - MIAI Institute (Efelia, ANR-22-CMAS-0005).


\printbibliography[title={References},heading=bibnumbered] 

\end{document}

\endinput 